\documentclass[preprint,12pt]{elsarticle}

\usepackage{graphicx,natbib}
\usepackage{amsmath,amssymb} 
\usepackage{color,bm,subfigure}
\usepackage{caption}
\usepackage{makecell}
\usepackage{multirow}
\usepackage{tabularx}

\journal{Pattern Recognition}

\begin{document}

\begin{frontmatter}



\title{Sparse, Collaborative, or Nonnegative Representation: Which Helps Pattern Classification?}

\author[a]{Jun Xu}
\author[b]{Wangpeng An}
\author[a]{Lei Zhang}
\author[a,c]{David Zhang}
\address[a]{Department of Computing, The Hong Kong Polytechnic University, Hong Kong, China}
\address[b]{Graduate School at Shenzhen, Tsinghua University, Shenzhen, China}
\address[c]{School of Science and Engineering, The Chinese University of Hong Kong (Shenzhen), Shenzhen, China}

\begin{abstract}
The use of sparse representation (SR) and collaborative representation (CR) for pattern classification has been widely studied in tasks such as face recognition and object categorization. Despite the success of SR/CR based classifiers, it is still arguable whether it is the $\ell_{1}$-norm sparsity or the $\ell_{2}$-norm collaborative property that brings the success of SR/CR based classification. In this paper, we investigate the use of nonnegative representation (NR) for pattern classification, which is largely ignored by previous work. Our analyses reveal that NR can boost the representation power of homogeneous samples while limiting the representation power of heterogeneous samples, making the representation sparse and discriminative simultaneously and thus providing a more effective solution to representation based classification than SR/CR. Our experiments demonstrate that the proposed NR based classifier (NRC) outperforms previous representation based classifiers. With deep features as inputs, it also achieves state-of-the-art performance on various visual classification tasks.  
\end{abstract}


\end{frontmatter}

\section{Introduction}
Pattern classification aims to find the correct class to which a query sample $\bm{y}$ belongs, given the training samples $\bm{X}$ from $K$ classes. In the past decades, many pattern classification algorithms \cite{lazebnik2006beyond,ScSPM,gao2010cvpr,dksvd,lcksvd,alexnet,he2016deep,PCANet,vapnik1998statistical,nsc,src,crc,croc,YANG20101454,YANG20121104,YANG2014535,procrc,FENG20132134,XIE2015447,LIN2018341} have been proposed. Among different types of pattern classification methods, one major category is the representation based methods \cite{vapnik1998statistical,nsc,src,crc,croc,procrc}. These methods first encode the query sample as a linear combination of the given training samples, and then assign the query sample to the corresponding class with the minimal distance or approximation error.\ One seminal work in this category is the Sparse Representation (SR) based Classifier (SRC)~\cite{src}, which enforces the $\ell_1$-norm sparsity~\cite{YANG20121104} on the coding vector of the query sample over all the training samples. Despite its success on face recognition, it is questioned that whether it is indeed the $\ell_1$-norm sparsity that makes SRC effective \cite{why2009,srreally,crc}. To this end, the Collaborative Representation (CR) based Classifier (CRC) \cite{crc} has been introduced to the pattern classification community, which reveals that it is the collaborative property induced by the $\ell_2$-norm that works for representation based pattern classification.

In SR/CR scheme based classifiers, the query sample is approximated by a linear combination of the training samples from all classes.\ Despite their success in face recognition, the representation based classifiers such as SRC \cite{src}, CRC \cite{crc}, and their extensions \cite{croc,procrc,WuJMM14,wu2014dcr} cannot explicitly avoid negative coding coefficients within complex optimization solutions.\ This is because there is no explicitly positive constraint on the coding coefficients of the query sample $\bm{y}$ over the training sample matrix $\bm{X}$.\ The negative coding coefficients indicate negative data correlations between the query sample and the training samples.\ This is to some extent counter-intuitive because the query sample $\bm{y}$ should be better approximated from the homogeneous samples to it with non-negative representation coefficients.\ The negative correlations are largely from the heterogeneity samples, which would bring little benefit to the prediction of the query sample.\ According to the viewpoint in \cite{nmfnature}, though mathematically feasible, it is not suitable to approximate the query sample by allowing the training samples to ``cancel each other out'' with additions and subtractions.\ Recently, researchers have attempted to explain the mechanism of SRC/CRC from the perspective of probabilistic subspace \cite{procrc}.\ However, its coding scheme still cannot avoid the negative coding coefficients on the approximation of the query sample.

As validated by \cite{belhumeur1997eigenfaces}, effective representation of the query sample should be dense over homogeneous data samples.\ This dense representation would be non-negative due to the homogeneity on these samples, reminiscent of non-negative matrix factorization (NMF) problem in machine learning community~\cite{nmfnature,Lee2000Algorithms}, which aims to approximate a data matrix via multiplication of two non-negative factorial matrices.\ Several NMF methods are proposed in the past few years~\cite{Ding2009Convex,Hoyer2004,CUI201838,ZDUNEK20082309,Babaee2015Discriminative}, trying to get more accurate approximation.\ However, NMF cannot be directly utilized for pattern classification.\ The key reason is that in this problem, the training samples need not to be fully non-negative.\ Hence, we only restrict the coding coefficients to be non-negative for representation based classification, which results in Non-negative Representation (NR).\ In fact, NR often leads to better performance over SR/CR for data representation \cite{nmfnature}.\ The main reason is that non-negative constraint in NR only allows non-negative combination of multiple training samples to reconstruct the query sample, which is compatible with the intuitive of combining parts into a whole \cite{nmfnature}.

\begin{figure}[t!]
\vspace{-0mm}
\centering
\subfigure{
\begin{minipage}[t]{0.2\textwidth}
\centering
\hspace{-5mm}
\raisebox{-0.15cm}{\includegraphics[width=1\textwidth]{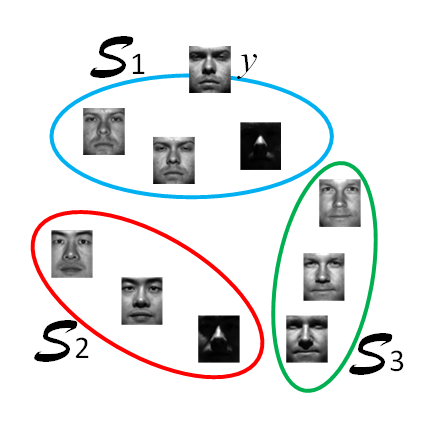}}
{(a)}
\end{minipage}
\begin{minipage}[t]{0.25\textwidth}
\centering
\hspace{-5mm}
\raisebox{-0.15cm}{\includegraphics[width=1\textwidth]{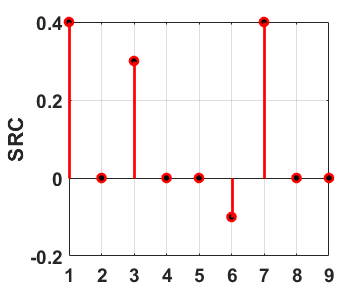}}
{(b)}
\end{minipage}
\begin{minipage}[t]{0.25\textwidth}
\centering
\hspace{-5mm}
\raisebox{-0.15cm}{\includegraphics[width=1\textwidth]{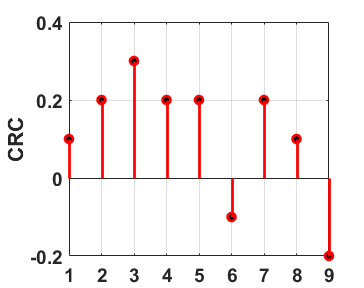}}
{(c)}
\end{minipage}
\begin{minipage}[t]{0.25\textwidth}
\centering
\hspace{-5mm}
\raisebox{-0.15cm}{\includegraphics[width=1\textwidth]{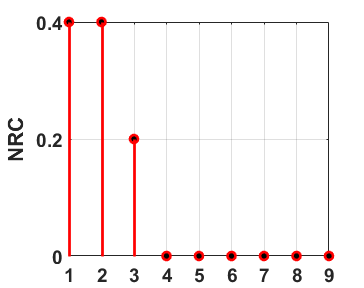}}
{(d)}
\end{minipage}
}\vspace{-4mm}
\caption{\small An illustrative comparison of SRC, CRC, and NRC. (a) A query sample $\bm{y}$ and 3 classes ($\mathcal{S}_{1}$, $\mathcal{S}_{2}$, and $\mathcal{S}_{3}$), each class having 3 training samples. Note that $\bm{y}$ is from class $\mathcal{S}_{1}$. (b)$\sim$(c) show the coding coefficients of $\bm{y}$ over the 9 training samples by SRC, CRC and NRC, respectively.}
\vspace{-4mm}
\label{f1}
\end{figure}

In this work, we investigate the power of Non-negative Representation (NR) for pattern classification tasks, such as face recognition, object recognition, and action recognition, etc. Motivated from the NMF problem, our major idea is that any given query sample $\bm{y}$ can be accurately coded by the non-negative coefficients over the homogeneous samples (i.e., samples from the same class with $\bm{y}$), which can determine the class label of $\bm{y}$.\ Constraining the coding coefficients to be non-negative can automatically boost the representational power of homogeneous samples while limiting the representation power of heterogeneous samples, making the representation sparse while discriminative.\ Based on above analysis, we propose a simple yet effective NR based Classifier (NRC) for pattern classification. Specifically, we utilize the non-negative constrained least square model \cite{bro1997fast,slawski2013}, which restricts the coding vector to be non-negative under the simple least square framework, for query sample encoding, and use the approximation residual from each class for classification. We solve the proposed NR model can be reformulated as a linear equality-constrained problem with two variables, and solved under the alternating direction method of multipliers framework \cite{admm}. Each variable can be solved efficiently in closed-form, and the convergence to the global optimum can be guaranteed. 

To validate our idea, we show an illustrative example in Fig. \ref{f1}. It can be seen that a query face sample $\bm{y}$ is encoded over 9 samples from 3 classes ($\mathcal{S}_{1}$, $\mathcal{S}_{2}$, and $\mathcal{S}_{3}$) in the YaleB dataset, 3 samples for each class. The query sample $\bm{y}$ is from class $\mathcal{S}_{1}$. By encoding $\bm{y}$ over the 9 samples, one can see that CRC generates dense coefficients, while SRC generates sparse coefficients but from all the 3 classes. In contrast, the coefficients generated by NRC are not only sparse but also from the same correct class. We also perform extensive comparison experiments on several benchmark datasets on face recognition, handwritten digit recognition, action recognition, object recognition, and fine grained visual classification tasks. The results demonstrate that the proposed NRC is very efficient, achieves higher accuracy than other SR/CR scheme based classifiers, and competing performance with state-of-the-art classifiers.

In summary, our contributions are as follows:
\begin{itemize}
\item We introduce the non-negative representation (NR) as a novel representation scheme;
	
\item We propose a novel NR based classifier (NRC) based on the simple non-negative least square model, which can be reformulated as a linear equality-constrained problem with two variables, and solved under the alternating direction method of multipliers framework \cite{admm};
	
\item We evaluate the proposed NRC classifier on several pattern recognition tasks, and demonstrate that NRC outperforms existing representation based approaches on various benchmark datasets.
\end{itemize}

The rest of this paper is organized as follows: We review the related work in Section 2. We present the proposed NRC based classifier in Section 3. In Section 4, we perform a comprehensive comparison on the proposed NRC classifier as well as several competing classifiers on various visual classification datasets. We conclude this paper in Section 5.

\section{Related Work}
\subsection{Representation based Pattern Classification}
The sparse representation based classifier (SRC) \cite{src} and collaborative representation based classifier (CRC) \cite{crc}, and their several extensions \cite{croc,procrc,WuJMM14,wu2014dcr} have been widely studied for various pattern classification tasks such as face recognition, handwritten digit recognition, object recognition, and action recognition, etc. Assume that we have $K$ classes of samples, denoted by $\{\bm{X}_{k}\},k=1,...,K$, where $\bm{X}_{k}$ is the sample matrix of class $k$. Each column of the matrix $\bm{X}_{k}$ is a training sample from the $k$-th class. The whole training sample matrix can be denoted as $\bm{X}=[\bm{X}_{1},...,\bm{X}_{K}]\in\mathbb{R}^{D\times N}$. Given a query sample $\bm{y}\in\mathbb{R}^{D}$, both SRC and CRC compute the coding vector $\bm{c}$ of $\bm{y}$ over $\bm{X}$ by solving the following minimization problem:
\vspace{-0mm}
\begin{equation}
\vspace{-0mm}
\label{e1}
\min_{\bm{c}}
\|\bm{y}-\bm{X}\bm{c}\|_{2}^{2}
+
\lambda
\|\bm{c}\|_{p}^{q},
\end{equation}
where $p=0$ or $1$, $q=1$ for SRC and $p=q=2$ for CRC, and $\lambda$ is the regularization parameter. The coding vector $\bm{c}$ can be written as $\bm{c}=[\bm{c}_{1}^{\top},..,\bm{c}_{K}^{\top}]^{\top}$, where $\bm{c}_{k}, k=1,...,K$ is the coding sub-vector of $\bm{y}$ over the sample sub-matrix $\bm{X}_{k}$.\ Assume that the query sample $\bm{y}$ belongs to the $k$-th class, then it is highly possible that $\bm{X}_{k}\bm{c}_{k}$ can give a good approximation of $\bm{y}$, i.e.,  $\bm{y}\approx\bm{X}_{k}\bm{c}_{k}$.\ Therefore, SRC and CRC perform classification by computing the approximation residual of $\bm{y}$ in each class as:
\vspace{-0mm}
\begin{equation}
\vspace{-0mm}
\label{e2}
\text{label}(\bm{y})
=
\arg\min_{k}\{\|\bm{y}-\bm{X}_{k}\bm{c}_{k}\|_{2}\}.
\end{equation}
The class (here $k$) with minimal residual would be the predicted class for $\bm{y}$. The major difference between SRC and CRC lies in the coding vector $\bm{c}$. For SRC, the coding vector $\bm{c}$ is very sparse induced by the $\ell_{1}$ norm, and the significant coefficients are mostly fall into $\bm{c}_k$. While for CRC, $\bm{c}$ is generally very dense over all classes, this is the collaborative property induced by the $\ell_{2}$ norm. The overall classification framework of the SRC/CRC classifiers is summarized in Algorithm 1.

\begin{table}[t!]
\vspace{-0mm}
\centering
\begin{tabular}{l}
\Xhline{1pt}
\textbf{Algorithm 1}: The SRC/CRC Algorithms
\\
\Xhline{0.5pt}
\textbf{Input}: Training sample matrix $\bm{X}=[\bm{X}_{1},...,\bm{X}_{K}]$, query sample $\bm{y}$;
\\
1. Normalize the columns of $\bm{X}$ to have unit $\ell_{2}$ norm;
\\
2. Compute the coding vector $\bm{\hat{c}}$ of $\bm{y}$ over $\bm{X}$ via 
\\
\qquad\qquad \qquad 
$\bm{\hat{c}}
=
\arg\min_{\bm{c}}
\|\bm{y}-\bm{X}\bm{c}\|_{2}^{2}
+
\lambda
\|\bm{c}\|_{p}^{q}
$;
\\
3. Compute the approximation residuals 
$
r_{k}=\|\bm{y}-\bm{X}_{k}\bm{\hat{c}}_{k}\|_{2}
$;
\\ 
\textbf{Output}:
Label of $\bm{y}$: 
$
\text{label}(\bm{y})
=
\arg\min_{k}\{r_{k}\}
$.
\\
\Xhline{1pt}
\end{tabular}
\vspace{-0mm}
\label{a1}
\end{table}

Due to efficient performance and elegant mathematical theory, CRC \cite{crc} has been extended in several directions. In~\cite{chi2014classification}, the authors solved the conventional multi-class classification
problem via a two-step framework, first finding the collaborative representation (CR) and then applying it to a multi-class classifier. Later, Timofte \textsl{et al.}~\cite{timofte2014adaptive} introduced a weighted CR classifier, which sets adaptive weights for different samples or features. In \cite{wu2015learned}, the authors recovered that it is still unclear that how to choose the weights and weights optimization in the method of \cite{timofte2014adaptive}. Thus, they proposed a learned collaborative representation based classifier. In \cite{procrc}, the authors proposed a probabilistic CR based classifier, which explains the working mechanism of SRC and CRC from the perspective of probability. 

In this work, we introduce a non-negative representation (NR) based scheme as an alternative to the widely studied SR/CR framework for pattern classification. The introduced NR scheme is inspired from the non-negative matrix factorization techniques~\cite{nmfnature,Lee2000Algorithms}, which is very different from the $\ell_{1}$ induced SR or the $\ell_{2}$ induced CR schemes.

\subsection{Non-negative Matrix Factorization}
Non-negative matrix factorization (NMF) is introduced by Daniel D.Lee and H.Sebastian Seung in~\cite{nmfnature,Lee2000Algorithms} for face representation, which consists in finding two non-negative factorial
matrices whose multiplication is an accurate approximation of the original sample matrix. Specifically, given the sample matrix $\bm{X}$, NMF aims to find two non-negative matrices $\bm{U}$ and $\bm{V}$ such that $\bm{X}\approx\bm{U}\bm{V}$. Here, $\bm{U}$ can be regarded as the base matrix while the $\bm{V}$ can be regarded as the encoding matrix. This is usually performed via 
\begin{equation}
\label{e3}
\min_{\bm{U}>0,\bm{V}>0}
\|
\bm{X}-\bm{U}\bm{V}
\|
_{F}^2,
\end{equation}
where $\|\cdot\|_{F}$ denotes the Frobenius norm. This problem is usually solved by alternating non-negative least squares method~\cite{Lee2000Algorithms}, in which $\bm{U}$ and $\bm{V}$ are iteratively updated by fixing the other one. Systematic study of the solution is not the theme of this paper, please read \cite{nmfreview} for comprehensive study.

The NMF representation should be able to discover the characteristics in local parts of the original sample matrix in an additive way. The
non-negative constraint in NMF only allows additive operation (without subtraction), and leads to the part-based representation~\cite{nmfnature,Lee2000Algorithms}. The two factorized non-negative matrices
are called basis matrix and coefficient matrix, respectively. NMF has recently been receiving increasing attention as a framework to serve numerous tasks such as face recognition \cite{li2001learning}, graph
clustering \cite{Graphnmf}, and document clustering \cite{xu2004document}, etc. However, the NMF framework~\cite{nmfnature,Lee2000Algorithms} also 
has its limitation~\cite{li2001learning,nmfreview}. One unavoidable issue of NMF is that it cannot explicitly produce sparse solutions \cite{nmfreview}. In \cite{zhang2011image}, the authors proposed to incorporate sparseness property into NMF and extended it to learn more specific features. Several other work have also been proposed~\cite{Ding2009Convex,Hoyer2004,CUI201838,ZDUNEK20082309,Babaee2015Discriminative} to extend NMF to broader applications. Different from these methods, in this work we only constraints the coefficients to be non-negative but allow the basis matrix to be the sample matrix, in which the samples could contain negative elements. The introduced non-negative representation based model can be solved very efficiently.

\section{Non-negative Representation based Classification}

\subsection{The Non-negative Representation Model}

The core idea of Sparse Representation based Classifier (SRC) and Collaborative Representation based Classifier (CRC) is to encode the query sample $\bm{y}$ over the whole training sample matrix $\bm{X}$ instead of each class-specific sample matrix $\bm{X}_k$, while the difference lies in that whether the constraint of $\ell_{1}$ norm or $\ell_{2}$ norm is imposed on the coding vector. However, SRC and CRC cannot avoid to generate negative coefficients in the coding vector. Though this is mathematically feasible, it is physically not suitable to reconstruct a sample by applying both additions and subtractions to training samples in real-world applications \cite{nmfnature}.

For representation based pattern classification task, the key problem is how to obtain a discriminative coding vector upon the training samples, which can be used to approximate the query sample from homogeneous training samples. Given a query sample $\bm{y}$ and the training sample matrix $\bm{X}$, the discriminative coding entries of $\bm{y}$ over $\bm{X}$ should be positive upon those homogeneous training samples of $\bm{y}$ from the same class, while be zero upon those heterogeneous training samples from different classes. This is based on the fact that the homogeneous samples are usually more similar to the query sample $\bm{y}$ with positive correlations. On the contrary, the heterogeneous samples should better contribute nothing to the approximation of the query sample. In SRC/CRC, since there is no restriction on the sign of the coding coefficients, the query sample $\bm{y}$ will be approximated by complex additions and subtractions of the training samples in $\bm{X}$ from different classes, which makes the physical explanation difficult. 

Based on the above discussion, we impose the non-negative constraint \cite{nngarrote,ds3} on the coding coefficients instead of the $\ell_{1}$-norm or $\ell_{2}$-norm regularizations. Given the query sample $\bm{y}\in\mathbb{R}^{D}$ and the training sample matrix $\bm{X}$ consisting of samples from several classes, i.e., $\bm{X}=[\bm{X}_{1},...,\bm{X}_{K}]\in\mathbb{R}^{D\times N}$, we employ the following non-negative representation (NR) based model to find the discriminative coding vector:
\vspace{-0mm}
\begin{equation}
\vspace{-0mm}
\begin{split}
\label{e4}
&
\min_{\bm{c}}
\|
\bm{y}
-
\bm{X}\bm{c}
\|_{2}^{2}
\quad 
\text{s.t.}
\quad
\bm{c}\ge0
.
\end{split}
\end{equation}
Here, we assume that there are $K$ classes of samples, denoted by $\{\bm{X}_{k}\},k=1,...,K$, where $\bm{X}_{k}$ is the sample matrix of class $k$. Each column of the matrix $\bm{X}_{k}$ is a training sample from the $k$-th class. 

Due to the non-negative constraint exposed on the coding vector $\bm{c}$, the NR model in Eq. (\ref{e4}) will select a few samples in $\bm{X}$ to approximate the query sample $\bm{y}$, naturally resulting in sparsity on $\bm{c}$.\ Meanwhile, it tends to find the homogeneous samples to represent $\bm{y}$ in order for accurate approximation, resulting in a discriminative representation.\ The ability to achieve sparsity and discriminability simultaneously makes NR a better choice than SR and CR for pattern classification.

\vspace{-0mm}
\subsection{Model Optimization}
\vspace{-0mm}

The proposed NR model is basically the non-negative least square (NNLS) \cite{bro1997fast,slawski2013} problem, which does not have a closed-form solution. To this end, we employ the variable splitting method \cite{courant1943,Eckstein1992} to solve the NR model (\ref{e4}). By introducing an auxiliary variable $\bm{z}$, we can reformulate the NR model into a linear equality-constrained problem with two variables $\bm{c}$ and $\bm{z}$:
\vspace{-0mm}
\begin{equation}
\vspace{-0mm}
\begin{split}
\label{e5}
&
\min_{\bm{c}}
\|
\bm{y}
-
\bm{X}\bm{c}
\|
_{2}^{2}
\quad 
\text{s.t.}
\quad
\bm{c}=\bm{z}
,
\bm{z}\ge0
.
\end{split}
\end{equation}
The problem (\ref{e5}) can be solved under the alternating direction method of multipliers (ADMM) \cite{admm} framework. The Lagrangian function of the problem (\ref{e5}) is
\begin{equation}
\begin{split}
\label{e6}
\mathcal{L}
(\bm{c},\bm{z},\bm{\delta},\rho)
=
&
\|
\bm{y}
-
\bm{X}\bm{c}
\|_{2}^{2}
+
\langle
\bm{\delta},\bm{z}-\bm{c}
\rangle
+
\frac{\rho}{2}
\|
\bm{z}
-
\bm{c}
\|_{2}^{2}
,
\end{split}
\end{equation}
where $\bm{\delta}$ is the augmented Lagrangian multiplier and $\rho>0$ is the penalty parameter. We initialize the vector variables $\bm{c}_{0}$, $\bm{z}_{0}$, and $\bm{\delta}_{0}$ to be zero vectors and set $\rho>0$ with a suitable value.\ Denote by ($\bm{c}_{t}$, $\bm{z}_{t}$) and $\bm{\delta}_{t}$ the optimization variables and the Lagrange multiplier at iteration $t$ ($t = 0, 1, 2, ...$), respectively.\ The variables can be updated by taking derivatives of the Lagrangian function (\ref{e6}) w.r.t. the variables $\bm{c}$ and $\bm{z}$ and setting the derivative function to be zero.
\vspace{1mm}
\\
(1) \textbf{Updating $\bm{c}$ while fixing $\bm{z}$ and $\bm{\delta}$}:
\vspace{-0mm}
\begin{equation}
\vspace{-0mm}
\begin{split}
\label{e7}
&
\min_{\bm{c}}
\|
\bm{y}
-
\bm{X}\bm{c}
\|_{2}^{2}
+
\frac{\rho}{2}
\|
\bm{c}
-
(
\bm{z}_{t}
+
\rho^{-1}
\bm{\delta}_{t}
)
\|_{2}^{2}
.
\end{split}
\end{equation}
This is a standard least squares regression problem with closed form solution:
\vspace{-0mm}
\begin{equation}
\vspace{-0mm}
\begin{split}
\label{e8}
\bm{c}_{t+1}
=
(\bm{X}^{\top}\bm{X}+\frac{\rho}{2}\bm{I})^{-1}
(\bm{X}^{\top}\bm{y}+\frac{\rho}{2}\bm{z}_{t}+\frac{1}{2}\bm{\delta}_{t})
\end{split}
\end{equation}
\vspace{0mm}
\\
(2) \textbf{Updating $\bm{z}$ while fixing $\bm{c}$ and $\bm{\delta}$}:
\vspace{-0mm}
\begin{equation}
\vspace{-0mm}
\begin{split}
\label{e9}
&
\min_{\bm{z}}
\|
\bm{z}
-
(
\bm{c}_{t+1}
-
\rho^{-1}
\bm{\delta}_{t}
)
\|_{2}^{2}
\quad 
\text{s.t.}
\quad
\bm{z}\ge0
.
\end{split}
\end{equation}
The solution of $\bm{z}$ is
\vspace{-0mm}
\begin{equation}
\vspace{-0mm}
\begin{split}
\label{e10}
\bm{z}_{t+1}=\max(0,\bm{c}_{t+1}-\rho^{-1}\bm{\delta}_{t}),
\end{split}
\end{equation}
where the ``$\max$'' operates element-wisely.
\vspace{1mm}
\\
(3) \textbf{Updating the Lagrangian multiplier $\bm{\delta}$}:
\vspace{-0mm}
\begin{equation}
\vspace{-0mm}
\begin{split}
\label{e11}
\bm{\delta}_{t+1}
&
=
\bm{\delta}_{t}
+
\rho
(\bm{z}_{t+1}-\bm{c}_{t+1})
.
\end{split}
\end{equation}
The above alternative updating steps are repeated until the convergence condition is satisfied or the number of iterations exceeds a preset threshold $T$.\ The convergence condition of the ADMM algorithm is: $\|\bm{c}_{t}-\bm{z}_{t}\|_{2}\le \text{Tol}$, $\|\bm{c}_{t+1}-\bm{c}_{t}\|_{2}\le \text{Tol}$, and $\|\bm{z}_{t+1}-\bm{z}_{t}\|_{2}\le\text{Tol}$ are simultaneously satisfied, where $\text{Tol}>0$ is a small tolerance value.\ Since the least square objective function and the linear equality and non-negative constraints are all strictly convex, the problem (\ref{e5}) solved by the ADMM algorithm is guaranteed to converge to a global optimal solution.\ We summarize the overall updating procedures in Algorithm 2.
\begin{table}[t!]
\vspace{-0mm}
\centering
\begin{tabular}{l}
\Xhline{1pt}
\textbf{Algorithm 2}: Solve the NR model (\ref{e3}) via ADMM
\\
\hline
\textbf{Input:} Query sample $\bm{y}$, training sample matrix $\bm{X}$, $\text{Tol}>0$, $T$, $\rho>0$;
\\
\textbf{Initialization:} $\bm{c}_{0}=\bm{z}_{0}=\bm{\delta}_{0}=\bm{0}$, $t=0$; 
\\
\textbf{for} $t=0:T-1$ \textbf{do}
\\
1. Update $\mathbf{c}_{t+1}$ as 
$
\bm{c}_{t+1}
=
(\bm{X}^{\top}\bm{X}+\frac{\rho}{2}\bm{I})^{-1}
(\bm{X}^{\top}\bm{y}+\frac{\rho}{2}\bm{z}_{t}+\frac{1}{2}\bm{\delta}_{t})
$;
\\
2. Update $\mathbf{z}_{t+1}$ as
$
\bm{z}_{t+1}=\max(0,\bm{c}_{t+1}-\rho^{-1}\bm{\delta}_{t})
$;
\\
3. Update $\bm{\delta}_{t+1}$ as
$
\bm{\delta}_{t+1}
=
\bm{\delta}_{t} + \rho(\bm{z}_{t+1}-\bm{c}_{t+1})
$;
\\
\quad \textbf{if} (Convergence condition is satisfied)
\\
4.\quad Stop;
\\
\quad \textbf{end if}
\\
\textbf{end for}
\\
\textbf{Output:} Coding vectors $\bm{z}_T$ and $\bm{c}_T$.
\\
\Xhline{1pt}
\end{tabular}
\label{a2}
\vspace{-0mm}
\end{table}

\textbf{Convergence Analysis}. The Algorithm 2 can be guaranteed to converge since the overall objective function (\ref{e5}) is convex with a global optimal solution. In Figure \ref{f1}, we show the convergence curve of the proposed NNLS algorithm by using the well-known MNIST dataset \cite{mnist}. It can seen that the maximal values in $|\bm{c}_{t+1}-\bm{c}_{t}|$, $|\bm{z}_{t+1}-\bm{z}_{t}|$, and $|\bm{c}_{t}-\bm{z}_{t}|$ approach to 0 simultaneously in 100 iterations.
\begin{figure}[t!]
\centering
\raisebox{-0.15cm}{\includegraphics[width=0.7\textwidth]{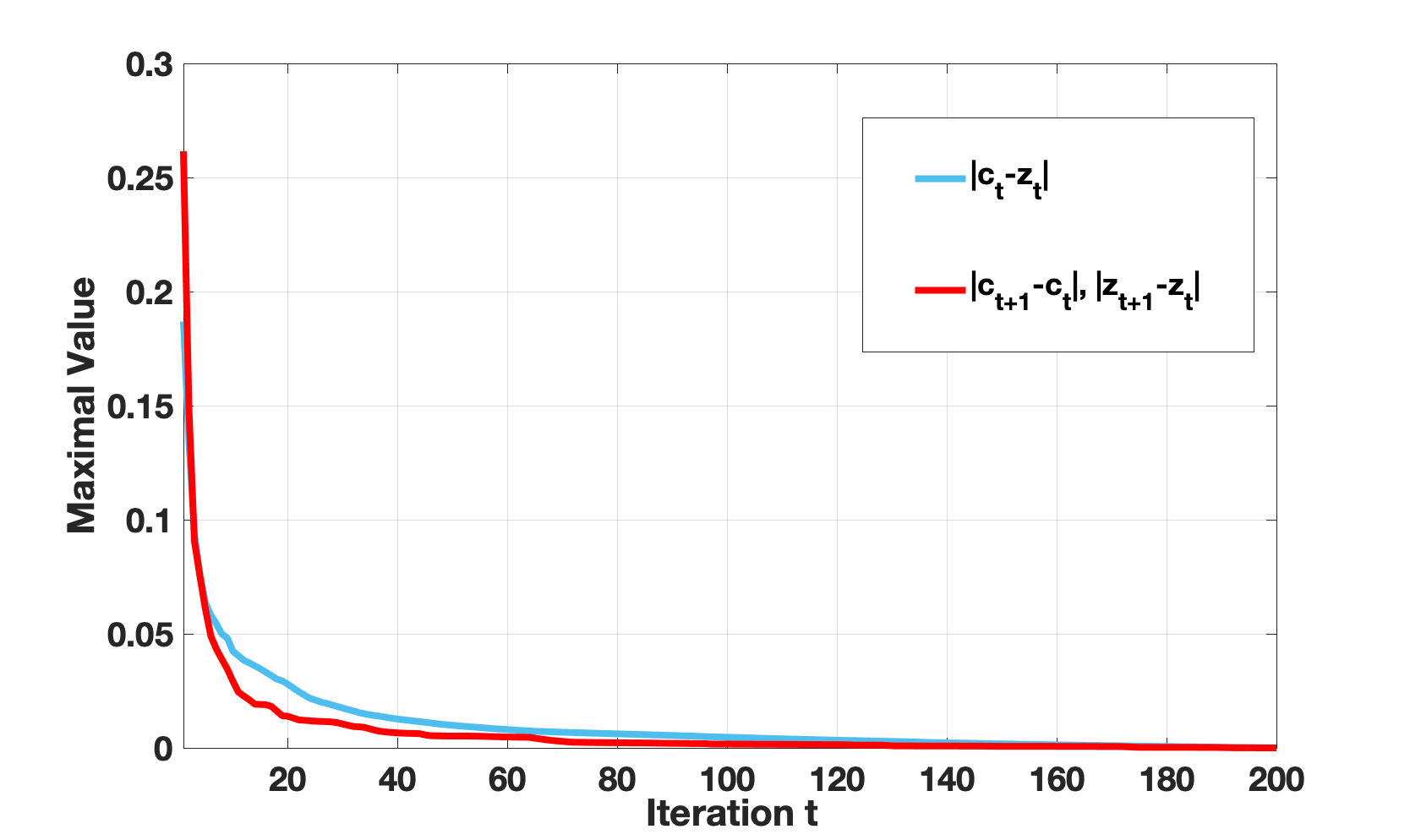}}
\caption{The convergence curves of maximal values in entries of $|\bm{c}_{t}-\bm{z}_{t}|$ (blue line), $|\bm{c}_{t+1}-\bm{c}_{t}|$ (red line), and $|\bm{z}_{t+1}-\bm{z}_{t}|$ (red line, roughly the same with $|\bm{c}_{t+1}-\bm{c}_{t}|$). The test set is the well-known MNIST dataset \cite{mnist}.}
\label{f1}
\end{figure}

\subsection{Classification}
Given the query sample $\bm{y}\in\mathbb{R}^{D}$ and the training sample matrix $\bm{X}=[\bm{X}_{1},...,\bm{X}_{K}]
\in\mathbb{R}^{D\times N}$, we first normalize $\bm{y}$ and each column of $\bm{X}$ to have unit $\ell_{2}$ norm, and then compute the coding vector $\hat{\bm{c}}$ of the query sample $\bm{y}$ over $\bm{X}$ via solving the problem (\ref{e4}). The classification of $\bm{y}$ by $\hat{\bm{c}}$ is similar to that of SRC/CRC (please refer to the Algorithm 1). We compute the class-specific representation residual $\|\bm{y}-\bm{X}_{k}\hat{\bm{c}}_{k}\|_{2}$, where $\hat{\bm{c}}_{k}$ corresponds to the coding sub-vector associated with class $k$. The proposed Non-negative Representation based Classifier (NRC) is summarized in Algorithm 3.

\begin{table}[t!]
\vspace{-0mm}
\centering
\begin{tabular}{l}
\Xhline{1pt}
\textbf{Algorithm 3}: The NRC Algorithm
\\
\Xhline{0.5pt}
\textbf{Input}: Training sample matrix $\bm{X}=[\bm{X}_{1},...,\bm{X}_{K}]$, query sample $\bm{y}$;
\\
1. Normalize the columns of $\bm{X}$ to have unit $\ell_{2}$ norm;
\\
2. Codes $\bm{y}$ over $\bm{X}$ via solving the NR model (\ref{e3}):
\\
\qquad\qquad
$
\hat{\bm{c}}
=
\arg\min_{\bm{c}}
\|
\bm{y}
-
\bm{X}\bm{c}
\|_{2}^{2}
\quad 
\text{s.t.}
\quad
\bm{c}\ge0
$;
\\
3. Compute the residuals
$
r_{k}=\|\bm{y}-\bm{X}_{k}\bm{\hat{c}}_{k}\|_{2}
$;
\\ 
\textbf{Output}:
$
\text{Label}(\bm{y})
=
\arg\min_{k}\{r_{k}\}
$.
\\
\Xhline{1pt}
\end{tabular}
\label{a3}
\vspace{-0mm}
\end{table}

\vspace{-0mm}
\subsection{Discussion}
\vspace{-0mm}

\textbf{Relation to other problems}. This work is inspired by the advantages of non-negative matrix factorization for parts-based facial analysis \cite{nmfnature}. The non-negativity property allows only the non-negative combination of multiple training samples in $\bm{X}$ to additively reconstruct the query sample $\bm{y}$, which is in accordance with the intuitive notion of ``combining parts into a whole" introduced in \cite{nmfnature}. Non-negativity is also in accordance with the biological modeling of visual data and often leads to better performance for data representation \cite{nmfnature}. In many real-world applications, the underlying signals represented by quantities can only take non-negative values by nature. Examples validating this point include pixel intensities in images, amounts of materials, chemical concentrations, and the compounds of end-members in hyper-spectral images, just to name a few. Even for realistic signals which may contain negative values, the representational coefficients are further required to be non-negative to avoid contradicting physical realities.

\textbf{Relation with NNLS}. Our NR model (\ref{e4}) shares the same mathematical formula with NNLS, which has been studied in \cite{slawski2013} for sparse recovery without regularization. The authors compared NNLS with the non-negative LASSO \cite{lasso}, and found that NNLS can achieve similar or even better performance on sparse recovery problems. Besides, the non-negative garrote has competitive performance with the subset selection method \cite{nngarrote}. The sparsity of NR can also be explained from the perspective of convex geometry. The sparsity of the non-negative constraints can be analyzed by studying the face lattice of the polyhedral cone generated by the columns of the training sample matrix \cite{nnls}.

\textbf{Sparsity and discriminative}. Like in SRC/CRC \cite{src,crc}, in NRC the query sample $\bm{y}$ is also coded over all the training samples. However, an evident distinction of NRC from SRC/CRC is that NRC restricts the coding coefficients to be non-negative and hence only the samples share similar structures (largely from the homogeneous class) to the query sample $\bm{y}$ can have positive coding coefficients and contribute positively to the approximation of $\bm{y}$. Seen in this light, the NRC based classifier can be guaranteed to perform well on representation based classification tasks. Overall, we argue that NRC brings discriminative property and sparsity simultaneously, and provides an ideal model for representation based pattern classification. 

\textbf{Perspective of convex hull}. In Eq. (\ref{e4}), the query sample $\bm{y}\in\mathbb{R}^{D}$ is consisted of non-negative combination of the homogeneous samples from $\bm{X}=[\bm{X}_{1},...,\bm{X}_{K}]\in\mathbb{R}^{D\times N}$, where $\bm{X}_{k}$ is the sample matrix including all the samples from the $k$-th class ($k=1,..,K$). Each sample in $\bm{X}_{k}$ can be viewed as a dictionary atom from the $k$-th class. Since the samples in $\bm{X}_{k}$ is very similar to each other, the linear combination of these dictionary atoms in $\bm{X}_{k}$ can be regarded as a convex hull. Different sample matrix form different convex hulls. Given the query sample $\bm{y}$, the non-negative constraint of $\bm{y}$ upon $\bm{X}$ will require all the reconstructed samples (or dictionary atoms) are in the interior of the correct convex hull. Hence, the non-negative constraint can induce correct representational coefficients from the correct sample matrix $\bm{X}_{k}$ for the query sample $\bm{y}$. Of course we can learn some discriminative yet compact dictionary atoms for each class as \cite{fddl,fddlijcv}, in this work, we employ the data samples directly as the dictionary atoms for simplicity.

\subsection{Complexity Analysis}

The Algorithms 2 and 3 may become very slow when the number of samples in $\bm{X}$ is much larger than the feature dimension, i.e., $N\gg D$. In order to make the proposed NRC scalable to large scale visual datasets, we employ the well-known Woodbury Identity Theorem \cite{higham2002accuracy} to reduce the computational cost for the inversion of the solution in Eq. (\ref{e7}). By this way, the update of $\bm{c}$ in (\ref{e7}) can be computed as follows:
\vspace{-0mm}
\begin{equation}
\vspace{-0mm}
\begin{split}
\label{e11}
\bm{c}_{t+1}
=
&
(
\frac{2}{\rho}\bm{I}
-
(\frac{2}{\rho})^{2}
\bm{X}^{\top}
(
\bm{I}
+
\frac{2}{\rho}\bm{X}\bm{X}^{\top}
)^{-1}
\bm{X}
)
(\bm{X}^{\top}\bm{y}
+
\frac{\rho}{2}
\bm{z}_{t}
+
\frac{1}{2}
\bm{\delta}_{t}
)
,
\end{split}
\end{equation}
which will save a lot of computational costs.

In Algorithm 2, the cost for updating $\bm{c}$ is $\mathcal{O}(DN^{2})$ due to the employment of Woodbury Identity Theorem. The cost for updating $\bm{z}$ is $\mathcal{O}(D)$. The costs for updating $\bm{\delta}$ is also $\mathcal{O}(D)$. So the overall complexity for Algorithm 2 is $\mathcal{O}(DN^{2}T)$, where $T$ is the number of iterations. In Algorithm 3 for pattern classification, the inversion in Eq.\ (\ref{e7}) is pre-stored and can be avoided in the testing stage. For each query sample $\bm{y}$, the costs for computing the residuals can be ignored. The overall cost for Algorithm 3 is $\mathcal{O}(N^{2}T)$.

\section{Experiments}

In this section, we compare the proposed NRC with state-of-the-art classifiers on various pattern classification tasks.

\subsection{Implementation Details}

The proposed NRC algorithm has two parameters: the iteration number $T$ and the penalty parameter $\rho$. With the increase of $T$, the accuracy will in general increase and finally converge to some rate.\ To balance accuracy and complexity, in all experiments we empirically set $T=5$ and determine $\rho$ by 5-fold cross validation on the training set.\ For the comparison methods, we use the source codes provided by the original authors, and tune their parameters to achieve their corresponding highest classification accuracies on different datasets.

\subsection{Comparison Methods and Datasets}

In Section 4.3, we first compare NRC with other representative and state-of-the-art representation based classifiers, including NSC \cite{nsc}, SRC \cite{src}, CRC \cite{crc}, CROC \cite{croc}, and ProCRC \cite{procrc}, on the classical face recognition datasets including the AR dataset \cite{ardatabase} and the Extended Yale B dataset \cite{YaleB}, and handwritten digit recognition datasets including the USPS dataset \cite{usps} and the MNIST dataset \cite{mnist}. The linear support vector machine (SVM) classifier \cite{liblinear} is also compared. In Section 4.4, we compare these competing methods for action recognition task on the Stanford 40 Actions dataset \cite{stanford40}, for object recognition task on the Caltech 256 dataset \cite{caltech256}, and for several challenging fine grained visual classification tasks on the Caltech-UCSD Birds-200-2011 dataset \cite{cubdataset}, the Oxford 102 Flowers dataset \cite{oxfordflower}, the Aircraft dataset \cite{fgvcaircraft}, and the Cars dataset \cite{fgvccars} datasets. We also compare the proposed method with state-of-the-art methods (such as FV-CNN \cite{fvcnn}, B-CNN \cite{bcnn}, and NAC \cite{nac}, etc.) on these fine grained visual classification datasets. 

\subsection{Experiments on Face and Handwritten Digit Classification}

\subsubsection{Face Recognition}
The \textbf{AR} dataset \cite{ardatabase} contains over 4,000 color images corresponding to 126 people's faces (70 men and 56 women) with different facial expressions, illumination conditions and occlusions. Following the settings in \cite{src,crc}, we choose a subset with only illumination and expression changes that contains 50 male subjects and 50 female subjects from the AR dataset in our experiments. For each subject, the seven images from Session 1 were used for training, with the other seven images from Session 2 for testing. The images are cropped to $60\times43$ and normalized to have unit $\ell_{2}$ norm. We project the images to a subspace of dimension $d=54,120,300$ by using PCA. The results of classification accuracy (\%) by the competing methods are listed in Table \ref{t1}. It can be seen that the proposed NRC achieves the highest accuracy on all dimensionalities. 

\begin{table}[ht!]
\caption{Classification accuracy (\%) of different algorithms on the AR dataset \cite{ardatabase}. The projected dimension by PCA is $d$.}
\label{t1}
\vspace{-5mm}
\begin{center}
\renewcommand\arraystretch{1}
\scriptsize
\begin{tabular}{|c||c|c|c|c|c|c|c|}
\Xhline{1pt}
$d$
&
{NSC} \cite{nsc}
&
{SVM} \cite{liblinear}
&
{SRC} \cite{src}
&
{CRC} \cite{crc}
&
{CROC} \cite{croc}
&
{ProCRC} \cite{procrc}
&
{NRC}
\\
\Xhline{1pt}
54
& 70.7 & 81.6 & 82.1 & 80.3 & 82.0 & 81.4 & \textbf{86.0}
\\
\hline
120
& 75.5 & 89.3 & 88.3 & 90.0 & 90.8 & 90.7 & \textbf{91.3} 
\\
\hline
300
& 76.1 & 91.6 & 90.3 & 93.7 & 93.7 & 93.7 & \textbf{94.0}
\\
\Xhline{1pt} 
\end{tabular}
\end{center}
\end{table}

The \textbf{Extended Yale B} dataset \cite{YaleB} has 2,432 face images from 38 human subjects, each having around 64 nearly frontal images taken under different illumination conditions. The original images are of $192\times168$ pixels. We resize the images to $54\times48$ pixels and normalize the images to have unit $\ell_{2}$ norm. As the experimental settings in \cite{crc}, we randomly split the dataset into two halves. Each half contains 32 images for each person. We use one half as the training samples, and the other half as the testing samples. We project the images to a subspace of dimension $d=84,150$, $300$ by PCA. The classification accuracies (\%) of the comparison methods are listed in Table \ref{t2}. One can see that the proposed NRC achieves almost the same performance as CROC, better than the other competing representation based classifiers.

\begin{table}[ht!]
\caption{Classification accuracy (\%) of different algorithms on the Extended Yale B dataset \cite{YaleB}. The results are averaged on 10 independent trials. The projected dimension by PCA is $d$.}
\label{t2}
\vspace{-5mm}
\begin{center}
\renewcommand\arraystretch{1}
\scriptsize
\begin{tabular}{|c||c|c|c|c|c|c|c|}
\Xhline{1pt}
$d$
&
{NSC} \cite{nsc}
&
{SVM} \cite{liblinear}
&
{SRC} \cite{src}
&
{CRC} \cite{crc}
&
{CROC} \cite{croc}
&
{ProCRC} \cite{procrc}
&
{NRC}
\\
\Xhline{1pt}
84
& 91.2 & 93.4 & 95.5 & 95.0 & 95.5 & 93.4 & \textbf{95.6}
\\
\hline
150
& 95.3 & 95.8 & 96.9 & 96.3 & \textbf{97.1} & 95.3 & \textbf{97.1 }
\\
\hline
300 
& 96.6 & 96.9 & 97.7 & 97.9 & \textbf{98.2} & 96.2 & \textbf{98.2}
\\
\Xhline{1pt} 
\end{tabular}
\end{center}
\end{table}

\subsubsection{Handwritten Digit Recognition}

The \textbf{USPS} dataset \cite{usps} contains 9,298 images for digit numbers from 0 to 9. The training set contains 7291 images while the testing set contains 2007 images. Each of the images in the USPS dataset is resized into $16\times16$ pixels. Similar to \cite{procrc}, we randomly select 50, 100, 200, and 300 images from each digit class of the training set as the training samples, and use all the samples in the testing set as the testing samples. We repeat the experiments on 10 independent trials and report the averaged results. We list the results of classification accuracy (\%) by different methods in Table \ref{t3}. It can be seen that the proposed NRC outperforms all the other competing methods on all cases. With the increasing of the number of training samples, the classification accuracy of all the comparison methods increases consistently, including the proposed NRC. However, one can see that the accuracy of NSC and ProCRC will not be further improved when the number of training samples increases from 200 to 300, while the proposed NRC can still increase from 94.6\% to 95.1\%.

\begin{table}[ht!]
\caption{Classification accuracy (\%) of different algorithms as a function of the number ($N$) of selected samples from each class for training on the USPS dataset \cite{usps}. The results are averaged on 10 independent trials.}
\label{t3}
\vspace{-5mm}
\begin{center}
\renewcommand\arraystretch{1}
\scriptsize
\begin{tabular}{|c||c|c|c|c|c|c|c|}
\Xhline{1pt}
Image \#
&
{NSC} \cite{nsc}
&
{SVM} \cite{liblinear}
&
{SRC} \cite{src}
&
{CRC} \cite{crc}
&
{CROC} \cite{croc}
&
{ProCRC} \cite{procrc}
&
{NRC}
\\
\Xhline{1pt}
$50$ 
& 91.2 & 91.6 & 91.4 & 89.2 & 91.9 & 90.9 & \textbf{92.3}
\\
\hline
$100$ 
& 92.2 & 92.5 & 93.1 & 90.6 & 91.3 & 91.9 & \textbf{93.7}
\\
\hline
$200$ 
& 92.8 & 93.1 & 94.2 & 91.4 & 91.7 & 92.2 & \textbf{94.6}
\\
\hline
$300$ 
& 92.8 & 93.2 & 94.8 & 91.5 & 91.8 & 92.2 & \textbf{95.1}
\\
\Xhline{1pt} 
\end{tabular}
\end{center}
\end{table}

The \textbf{MNIST} dataset \cite{mnist} contains 60,000 training images and 10,000 testing images for digit numbers from 0 to 9. The original images are of size $28\times28$. As in \cite{procrc}, we resize each image into $16\times16$ pixels. We randomly selected 50, 100, 300, and 600 samples from each class of the training set for training, and use all the samples in the testing set for testing. Different from the experimental settings in \cite{procrc}, for each image in the MNIST dataset, we extract its feature vector by using the scattering convolution network (SCN) \cite{scn}. The feature vector is a concatenation of coefficients in each layer of the network, and is translation invariant and deformation stable. Each feature vector is of length 3,472. The feature vectors for all images are then projected into a subspace of dimension 500 using PCA. We run the experiments for 10 independent trials and report the averaged classification accuracy (\%). The results are listed in Table \ref{t4}. It can be seen that the proposed NRC outperforms the comparison methods no matter how many images (50, 100, 200, or 600) from each class are chosen as the training samples. With the increase of the number of training samples, the classification accuracy of all the competing methods increases consistently. Since stronger features are extracted by using SCN \cite{scn}, the classification accuracies of all comparison methods are higher than the corresponding results reported in \cite{procrc}.


\begin{table}[ht!]
\caption{Classification accuracy (\%) of different algorithms as a function of the number ($N$) of selected samples from each class for training on the MNIST dataset \cite{mnist}. The results are averaged on 10 independent trials. The samples are projected onto a 500-dimensional space by PCA.}
\label{t4}
\vspace{-5mm}
\begin{center}
\renewcommand\arraystretch{1}
\scriptsize
\begin{tabular}{|c||c|c|c|c|c|c|c|}
\Xhline{1pt}
Image \#
&
{NSC} \cite{nsc}
&
{SVM} \cite{liblinear}
&
{SRC} \cite{src}
&
{CRC} \cite{crc}
&
{CROC} \cite{croc}
&
{ProCRC} \cite{procrc}
&
{NRC}
\\
\Xhline{1pt}
$50$ 
& 97.3 & 97.4 & 95.6 & 97.4 & 97.0 & 97.2 & \textbf{97.8}
\\
\hline
$100$ 
& 97.8 & 98.1 & 96.8 & 98.3 & 98.3 & 98.0 & \textbf{98.3}
\\
\hline
$300$ 
& 98.5 & 98.5 & 97.9 & 98.7 & 98.7 & 98.5 & \textbf{98.8}
\\
\hline
$600$ 
& 98.6 & 98.6 & 98.0 & 98.8 & 98.8 & 98.6 & \textbf{99.0} 
\\
\Xhline{1pt}
\end{tabular}
\end{center}
\end{table}

\subsubsection{Comparison on Speed}

We compare the running time (in second) of the proposed NRC and the competing representation based classifiers by processing one test image on the {MNIST} dataset \cite{mnist} as described in Section 4.3.2. We randomly selected 300 samples from each class of the training set for training, and use all the samples in the testing set for testing. The features extracted by SCN \cite{scn} are projected onto a 500-dimension subspace by PCA. All experiments are run under the Matlab environment and on a machine with 3.50 GHz CPU and 32 GB RAM. Table \ref{t5} lists the running time of different methods. The ProCRC and CRC methods have closed-form solutions and have the same speed, which is faster than CROC and much faster than SRC. The proposed NRC algorithm need several iterations in the ADMM algorithm, and hence is a little slower than CRC and ProCRC, but still faster than SRC and CROC.

\begin{table}[ht!]
\caption{Running time (second) of different methods on the {MNIST} dataset \cite{mnist}. The samples are projected onto a 500-dimensional space by PCA.}
\label{t5}
\scriptsize
\vspace{-5mm}
\begin{center}
\renewcommand\arraystretch{1}
\scriptsize
\begin{tabular}{|c||c|c|c|c|c|c|c|c|}
\Xhline{1pt}
Methods
&
{Softmax}
&
{NSC} 
&
{SVM} 
&
{SRC} 
&
CRC
&
{CROC}
&
{ProCRC}
&
{NRC}
\\
\hline
Time (s)
& 0.03 & 0.04 & 0.03 & 0.43 & 0.07 & 0.22 & 0.07 & 0.15
\\
\Xhline{1pt}
\end{tabular}
\end{center}
\end{table}

\subsection{Experiments on Large Scale Pattern Classification}

To take more comprehensive evaluation of NRC on other pattern classification tasks, we compare it with the state-of-the-art methods on six challenging pattern classification datasets: the Stanford 40 Actions dataset \cite{stanford40} for action recognition, the Caltech-256 dataset \cite{caltech256} for large scale object recognition, and the Caltech-UCSD Birds (CUB200-2011) dataset \cite{cubdataset}, the Oxford 102 Flowers dataset \cite{oxfordflower}, the Aircraft dataset \cite{fgvcaircraft}, the Cars dataset \cite{fgvccars} for fine-grained object recognition.

\subsubsection{Datasets and Settings}

The \textbf{Stanford 40 Actions} dataset \cite{stanford40} contains 9,352 images of 40 different classes of human actions with $180\sim300$ images per action, e.g., brushing teeth, reading book, and cleaning the floor, etc. Similar to the experimental settings in \cite{procrc}, we follow the training-testing split settings suggested in \cite{stanford40}, and randomly choose 100 images from each class as the training samples and use the remaining images as the testing samples.

The \textbf{Caltech-256} dataset \cite{caltech256} is consisted of 256 categories of objects, in which there are at least 80 images for each category. This large dataset has a total number of 30,608 images. We randomly choose 30 images from each object category as the training samples, and use the remaining images as the testing samples. For fair comparison, we run the proposed NRC and the comparison methods for 10 independent trials for each partition and report the averaged classification accuracy results (\%).

The \textbf{Caltech-UCSD Birds (CUB200-2011)} dataset \cite{cubdataset} contains 11,788 bird images, which is a widely-used benchmark for fine-grained visual classification. There are totally 200 bird species and the number of images per specie is around 60. The significant variations in pose, viewpoint, and illumination inside each class make this dataset very challenging for visual classification. We adopt the publicly available split \cite{cubdataset,procrc}, which use nearly half of the images in this dataset as the training samples and the other half as the testing samples.

The \textbf{Oxford 102 Flowers} dataset \cite{oxfordflower} is another fine-grained visual classification benchmark which contains 8,189 images from 102 flower classes. Each class contains at least 40 images, in which the flowers appear at different scales, pose, and lighting conditions. This dataset is very challenging since there exist large variations within the same class but small variations across different classes \cite{procrc}.

The \textbf{Aircraft} dataset \cite{fgvcaircraft} contains images of 100 different aircraft model variants, and there are 100 images for each model. The aircrafts appear at different scales, design structures, and appearances, making this dataset very difficult for visual classification task.\ We adopt the training-testing split protocol provided by \cite{fgvcaircraft} to design our experiments.

The \textbf{Cars} dataset \cite{fgvccars} is consisted of 16,185 images from 196 car classes. Each class has around 80 images with different sizes and heavy clutter background, which makes this dataset challenging for pattern classification. We use the same split scheme provided by \cite{fgvccars}, in which 8,144 images are employed as the training samples and the other 8,041 images are employed as the testing samples.

Following the experimental setting in \cite{procrc}, on the Stanford 40 Actions dataset, the Caltech-256 dataset, the Caltech-UCSD Birds (CUB200-2011) dataset and the Oxford 102 Flowers dataset, we employ two different types of features to demonstrate the effectiveness of the proposed NRC. For the first type features, we employ the VLFeat library \cite{vedaldi2010vlfeat} to extract the Bag-of-Words feature based on SIFT \cite{sift} (refer to as BOW-SIFT feature). The size of patch and stride are set as $16\times16$ and 8 pixels, respectively. The codebook is trained by the k-means algorithm, and we use a 2-level spatial pyramid representation \cite{lazebnik2006beyond}. The final feature dimension of each image is 5,120. For the second type features, we use VGG-verydeep-19 \cite{vggnet} to extract CNN features (refer to VGG19 features). We use the activations of the penultimate layer as local features, which are extracted on 5 scales $\{2^{s}, s = -1, -0.5, 0, 0.5, 1\}$. We pool all local features together regardless of scales and locations. The final feature dimension of each image is 4,096 for all datasets. The BOW-SIFT and VGG19 features are both $\ell_{2}$ normalized.\ On the Aircraft and Cars datasets, we use the CNN features extracted by a VGG-16 network \cite{vggnet} in the experiments.

\subsubsection{Evaluation of Representation based Classifiers with BOW-SIFT and VGG Features}

The accuracies (\%) by different methods on the six datasets are listed in Table \ref{t6}. As we can see, by using the same BoW-SIFT or VGG features, the proposed NRC achieves consistently higher accuracies than the previous representation based classifiers, including  SRC \cite{src}, CRC \cite{crc}, and their extensions CROC and ProCRC \cite{croc,procrc}. Specifically, with the VGG19 features, NRC achieves performance gains of at least 1.0\%, 1.1\%, 0.8\% and 0.5\% over the other classifiers on the Stanford 40 Actions, the Caltech-256, the CUB200-2011, and the Flower 102 datasets, respectively. With the BOW-SIFT features, the corresponding performance gains of NRC over the other classifiers are at least 0.6\%, 0.9\%, 0.3\%, and 3.2\%, respectively. With the VGG16 features, the performance gains of NRC over the other classifiers are at least 0.6\% and 0.5\% on the Aircraft and the Cars datasets, respectively. This demonstrates that NR is indeed more effective than previous SR and CR based classification methods.

\begin{table*}[ht!]
\caption{Classification accuracy (\%) of different classifiers on the above mentioned six datasets with BoW-SIFT or VGG features.}
\vspace{-5mm}
\label{t6}
\label{taba}
\begin{center}
\renewcommand\arraystretch{1}
\footnotesize
\begin{tabular}{|c||c|c|c|c|c|c|c|c|c|c|}
\hline
\multirow{2}{*}{Methods}
&
\multicolumn{2}{c|}{Standford 40}
&
\multicolumn{2}{c|}{Caltech 256}
&
\multicolumn{2}{c|}{CUB200-2011}
&
\multicolumn{2}{c|}{Flower 102}
&
\multicolumn{1}{c|}{Aircraft}
&
\multicolumn{1}{c|}{Cars}
\\
\cline{2-11} 
&
\scriptsize{SIFT}
&
\scriptsize{VGG19}
&
\scriptsize{SIFT}
&
\scriptsize{VGG19}
&
\scriptsize{SIFT}
&
\scriptsize{VGG19}
&
\scriptsize{SIFT}
&
\scriptsize{VGG19}
&
\scriptsize{VGG16}
&
\scriptsize{VGG16}
\\
\hline
\hline
Softmax & 21.1 & 77.2 & 25.8 & 75.3 & 8.2 & 72.1 & 46.5 & 87.3 & 85.6 & 88.7
\\
\hline
NSC \cite{nsc} & 22.1 & 74.7 & 25.8 & 80.2 & 8.4 & 74.5 & 46.7 & 90.1 & 85.5 & 88.3
\\
\hline 
SRC \cite{src}& 24.2 & 78.7 & 26.9 & 81.3 & 7.7 & 76.0 & 47.2 & 93.2 & 86.1 & 89.2
\\
\hline
CRC \cite{crc}& 24.6 & 78.2 & 27.4 & 81.1 & 9.4 & 76.2 & 49.9 & 93.0 & 86.7 & 90.0
\\
\hline
CROC \cite{croc}& 24.5 & 79.2 & 27.9 & 81.7 & 9.1 & 76.2 & 49.4 & 93.1 & 86.9 & 90.3
\\
\hline
ProCRC \cite{procrc}& 28.4 & 80.9 & 29.6 & 83.3 & 9.9 & 78.3 & 51.2 & 94.8 & 86.8 & 90.1
\\
\hline
NRC & \textbf{29.0} & \textbf{81.9} & \textbf{30.5} & \textbf{84.4} & \textbf{10.2} & \textbf{79.1} & \textbf{54.4} & \textbf{95.3} & \textbf{87.5} & \textbf{90.8}
\\
\hline
\end{tabular}
\end{center}
\vspace{-6mm}
\end{table*}

\subsubsection{Comparison with State-of-the-art Methods}
In this section, we compare NRC, by employing the VGG19 or VGG16 features, with other representative and state-of-the-art methods on the six datasets described in Section 4.4.1. Note that the compared CNN based methods use even stronger network architectures or features than our employed VGG19 features.

On the Stanford 40 Actions dataset, we compare the proposed NRC with the state-of-the-art methods such as AlexNet network \cite{alexnet}, VGG19 network \cite{vggnet}, EPM \cite{epm}, and ASPD \cite{aspd2015}.\ Note that EPM and ASPD are the leading methods on action recognition in still images.\ The classification accuracies are listed in Table \ref{t7}. One can see that NRC outperforms over the other methods by an improvement of at least 4.7\%.

\begin{table*}[ht!]
\caption{Classification accuracy (\%) of different methods on the Standford 40 datasets.}
\vspace{-5mm}
\label{t7}
\label{taba}
\begin{center}
\renewcommand\arraystretch{1.0}
\scriptsize
\begin{tabular}{|c||c|c|c|c|c|c|c|}
\hline
Methods
&
AlexNet \cite{alexnet}
&
EPM \cite{epm}
&
ASPD \cite{aspd2015}
&
VGG19 \cite{vggnet}
&
NRC
\\
\cline{1-6} 
Accuracy & 68.6 & 72.3 & 75.4 & 77.2 & \textbf{81.9}
\\
\hline
\end{tabular}
\end{center}
\end{table*}

On the Caltech-256 dataset, we compare the proposed NRC with the state-of-the-art methods such as AlexNet network \cite{alexnet}, VGG19 network \cite{vggnet}, M-HMP \cite{mhmp}, CNN-S \cite{Chatfield14}, ZF \cite{zf2014}, NAC \cite{nac}, and SJFT \cite{sjft}.\ These methods are the leading players among the others on large scale pattern classification tasks.\ Note that SJFT \cite{sjft} is a CNN based method employing transfering learning scheme to borrow treasures from abundant training data of other domains.\ The results are listed in Table \ref{t8}, one can see that NRC outperforms over other methods by an improvement of at least 0.6\%.

\begin{table*}[ht!]
\caption{Classification accuracy (\%) of different methods on the Caltech-256 datasets.}
\vspace{-5mm}
\label{t8}
\label{taba}
\begin{center}
\renewcommand\arraystretch{1.0}
\scriptsize
\begin{tabular}{|c||c|c|c|c|c|c|c|c|c|}
\hline
Methods
&
{M-HMP}
&
{ZF}
&
AlexNet 
&
CNN-S 
&
NAC 
&
VGG19 
&
SJFT
&
{NRC}
\\
\cline{1-9} 
Accuracy & 50.7 & 70.6 & 71.1 & 74.2 & 81.4 & 82.3 & 83.8 & \textbf{84.4} 
\\
\hline
\end{tabular}
\end{center}
\end{table*}

On the CUB200-2011 dataset, we compare the proposed NRC with the state-of-the-art methods such as AlexNet network \cite{alexnet}, VGG19 network \cite{vggnet}, POOF \cite{poof}, FV-CNN \cite{fvcnn}, PN-CNN \cite{pncnn}, and NAC \cite{nac}.\ Note that NAC is a part based method which employ data augmentation techniques, and achieve state-of-the-art accuracy on fine-grained recognition tasks.\ The accuracy results are listed in Table \ref{t9}.\ One can see that NRC is only slightly  inferior to NAC but still outperforms the other methods.
\begin{table*}[ht!]
\caption{Classification accuracy (\%) of different methods on the CUB200-2011 datasets.}
\vspace{-5mm}
\label{t9}
\label{taba}
\begin{center}
\renewcommand\arraystretch{1.0}
\scriptsize
\begin{tabular}{|c||c|c|c|c|c|c|c|}
\hline
Methods
&
AlexNet
&
{POOF}
&
{FV-CNN}
&
VGG19
&
PN-CNN
&
{NAC}
&
{NRC}
\\
\cline{1-8} 
Accuracy & 52.2 & 56.9 & 66.7 & 71.9 & 75.7 & \textbf{81.0} & 79.1 
\\
\hline
\end{tabular}
\end{center}
\end{table*}

On the Flower 102 dataset, we compare the proposed NRC with the state-of-the-art methods such as AlexNet network \cite{alexnet}, VGG19 network \cite{vggnet}, BiCoS \cite{bicos}, DAS \cite{das}, GMP \cite{gmp}, OverFeat \cite{OverFeat}, and NAC \cite{nac}.\ The results are listed in Table \ref{t10}.\ One can see that NRC achieves similar accuracy with NAC and at least 2.2\% performance gain than the other methods.

\begin{table*}[ht!]
\caption{Classification accuracy (\%) of different methods on the Flower 102 datasets with VGG19 features.}
\vspace{-5mm}
\label{t10}
\label{taba}
\begin{center}
\renewcommand\arraystretch{1.0}
\scriptsize
\begin{tabular}{|c||c|c|c|c|c|c|c|c|c|}
\hline
Methods
&
BiCoS
&
DAS 
&
GMP
&
OverFeat
&
AlexNet
&
VGG19
&
NAC
&
{NRC}
\\
\cline{1-9} 
Accuracy & 79.4 & 80.7 & 84.6 & 86.8 & 90.4 & 93.1 & \textbf{95.3} & \textbf{95.3}
\\
\hline
\end{tabular}
\end{center}
\end{table*}

For the Aircraft dataset and Cars dataset, we compare the proposed NRC with state-of-the-art VGG19 network \cite{vggnet}, Symbiotic \cite{symbiotic}, FV-FGC \cite{fvfgc}, B-CNN method \cite{bcnn}.\ Note that B-CNN achieves state-of-the-art performance on these two datasets.\ The results on classification accuracy (\%) are listed in Table \ref{t11}.\ One can see that, on the Aircraft dataset, the proposed NRC achieves an improvement of 3.4\% over the B-CNN. On the Cars dataset, the proposed NRC achieves an improvement of 0.2\% over the B-CNN.\
\begin{table}[ht!]
\caption{Classification accuracy (\%) of different methods on the Aircraft and Cars datasets with VGG16 features.}
\label{t11}
\vspace{-5mm}
\begin{center}
\renewcommand\arraystretch{1}
\footnotesize
\begin{tabular}{|c||c|c|c|c|c|}
\Xhline{1pt}
Dataset
&
VGG16 \cite{vggnet}
&
{Symbiotic} \cite{symbiotic}
&
{FV-FGC} \cite{fvfgc}
&
{B-CNN} \cite{bcnn}
&
{NRC}
\\
\hline
Aircraft
& 85.6 & 72.5 & 80.7 & 84.1 & \textbf{87.5}
\\
\hline
Cars
& 88.7 & 78.0 & 82.7 & 90.6 & \textbf{90.8}
\\
\Xhline{1pt}
\end{tabular}
\end{center}
\end{table}

All the above results demonstrate that, with deep CNN features, the proposed NRC can achieve comparable or even better performance than state-of-the-art methods on various visual classification datasets. It is a general and effective visual classifier.

\section{Conclusion}

Sparse representation based classifier (SRC) and collaborative representation based classifier (CRC) have been well-studied for face recognition. In this work, we provided an alternative classifier for pattern classification. Specifically, we investigated the attractive property of non-negative representation (NR) for pattern classification. We revealed that NR can enhance the representational power of homogeneous samples while limiting the representational power of heterogeneous samples, naturally resulting a sparse while discriminative encoding for predicting the query sample. Based on the NR scheme, we proposed a NR based model for classification. We solve the model via variable splitting techniques under the alternating direction method of multipliers framework. Based on the NR based model, we proposed a non-negative representation based classifier (NRC). Extensive experiments on various visual classification datasets validated the effectiveness of the proposed NRC classifier, and demonstrated that NRC outperforms previous representation based classifiers such as SRC/CRC. Besides, with deep features as inputs, NRC also achieves state-of-the-art performance on various large scale pattern classification tasks.

\section{Reference}
\scriptsize
\bibliographystyle{unsrt}
\bibliography{egbib}

\end{document}